\pdfoutput=1
\documentclass[conference]{IEEEtran}
\IEEEoverridecommandlockouts
\usepackage{cite}
\usepackage{amsmath,amssymb,amsfonts}
\usepackage{algorithmic}
\usepackage{subfigure,bm,balance}
\usepackage{graphicx}
\usepackage{textcomp}
\usepackage{xcolor}
\usepackage{hyperref}

\begin{document} 
\title{FuSSI-Net: Fusion of Spatio-temporal Skeletons for Intention Prediction Network}
\author{Francesco Piccoli$\textsuperscript{1*}$, Rajarathnam Balakrishnan$\textsuperscript{1*}$, Maria Jesus Perez$\textsuperscript{1*}$, Moraldeepsingh Sachdeo$\textsuperscript{1*}$, Carlos Nuñez$\textsuperscript{1*}$,\\ 
Matthew Tang$\textsuperscript{1*}$, Kajsa Andreasson$\textsuperscript{2*}$,Kalle Bjurek$\textsuperscript{2*}$, Ria Dass Raj$\textsuperscript{2*}$, Ebba Davidsson$\textsuperscript{2*}$, \\
Colin Eriksson$\textsuperscript{2*}$, Victor Hagman$\textsuperscript{2*}$, Jonas Sjöberg$\textsuperscript{2}$, Ying Li $\textsuperscript{3}$, 
L. Srikar Muppirisetty$\textsuperscript{4}$, Sohini Roychowdhury$\textsuperscript{3}$ \\
$\textsuperscript{1}$ University of California, Berkeley, CA 94720, USA\\
$\textsuperscript{2}$ Chalmers University of Technology, Department of Electrical Engineering, Göteborg, Sweden\\
$\textsuperscript{3}$ Volvo Cars Technology USA, Mountain View, CA-94043
$\textsuperscript{4}$ Volvo Car Corporation, SE-405 31 Göteborg, Sweden\\
\thanks{* All student authors have equal contribution.}
 \vspace{-0.5cm}}
\maketitle

\begin{abstract}
 Pedestrian intention recognition is very important to develop robust and safe  autonomous  driving  (AD) and  advanced  driver  assistance  systems  (ADAS) functionalities for urban driving. In this work, we develop an end-to-end pedestrian intention framework that performs well on day- and night- time scenarios. Our framework relies on objection detection bounding boxes combined with skeletal features of human pose. We study early, late, and combined (early and late) fusion mechanisms to exploit the skeletal features and reduce false positives as well to improve the intention prediction performance. The early  fusion  mechanism  results in AP of 0.89 and precision/recall of 0.79/0.89 for pedestrian intention classification. 
 Furthermore, we propose three new metrics to properly evaluate the pedestrian intention systems. Under these new evaluation metrics for the intention prediction, the proposed  end-to-end network offers accurate pedestrian intention up to half a second ahead of the actual \textit{risky maneuver}. 
\end{abstract}


\section{Introduction}
\begin{figure}[ht!]
    \centering
    \includegraphics[width=\columnwidth]{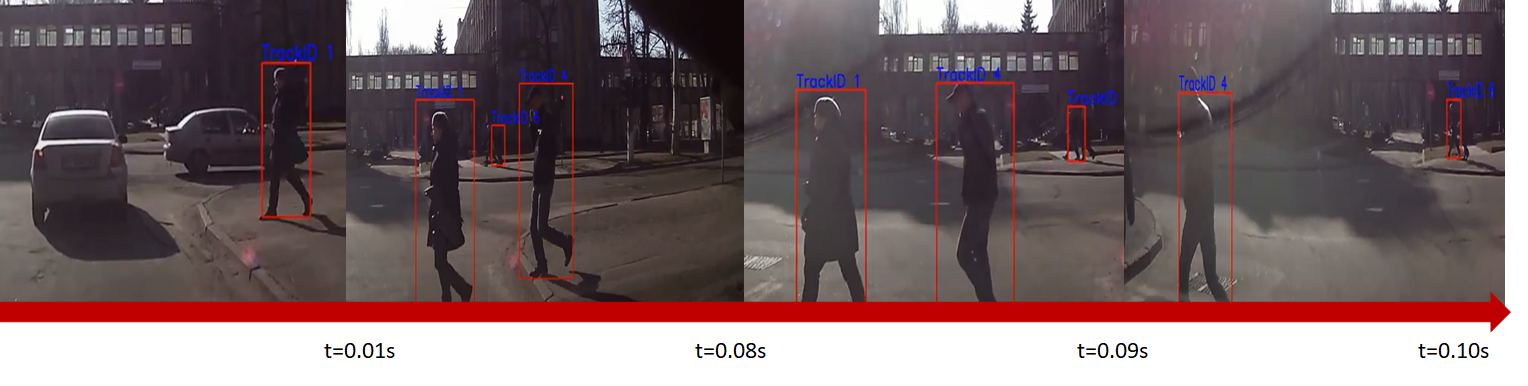}
    \caption{Examples of pedestrian intention prediction with respect to the ego-vehicle. Red bounding boxes predicted over a time sequence represent pedestrians that pose risk to the vehicle.
}
    \label{seq}
    \vspace{-0.4cm}
\end{figure}

Active safety functionalities for autonomous driving (AD) and advanced driver assistance systems (ADAS) in urban scenarios rely heavily on smart detection of the ego-vehicle environment conditions  to enhance people's safety \cite{varytimidis}. While intentions of other visible surrounding vehicles on the road can still be predicted through indicator/blinker signals, accurate detection and prediction of pedestrian and bicyclist intentions still remains a challenge at road cross  sections  \cite{rehder,varytimidis,saleh2}. Pedestrian intention prediction refers to automatically estimating the positions and intentions of pedestrians in the following few seconds with the goal of evaluating the individual risk associated with respect to the ego vehicle \cite{saleh2}. The information regarding the relative position of every other road user in future time frames with respect to the ego-vehicle is essential for lowering the false positive rates of collision avoidance alerts and systems. An example of detected risky pedestrians to ego-vehicles is shown in Fig. \ref{seq}. A smart collision avoidance system that detects pedestrian intention to cross or not can be significantly useful for anticipating potential risk posed to the ego-vehicle by pedestrians. 

With the surge in object detection and tracking algorithms over the past few years, there have been several works that have been directed towards designing modules towards pedestrian intention and path prediction.
In \cite{varytimidis}, the head orientation of pedestrians and their corresponding motion are detected by zooming into the regions corresponding to the head and legs, followed by oriented gradients and local binary pattern features for classification if a pestrian is crossing or not, using support vector machines or convolutional neural networks (CNNs).  The work \cite{rehder} implements a destination prediction network that is trained using CNN and a long short term memory (LSTM) model followed by a topology and planning network that utilize environmental features. While this work significantly differs from other object detection-based methods, it relies on locally curated datasets for performance analysis.  

The work in \cite{fang} takes a different approach for pedestrian and bicyclist detection than standard CNN modules. Here, a 9-point skeleton system is fitted for each pedestrian followed by crossing vs not-crossing classification. This work has shown state-of-the-art performances and hence we benchmark this method in this paper. Other work in \cite{ying} only focuses on tracking, so it cannot perform the intention prediction.

In spite of the existing works so far, there continues to be a need for an accurate end-to-end pedestrian intention prediction system that can utilize predicted future locations of pedestrians for vehicle ego-motion and path planning. There is a need for leveraging the advantages of various object detection systems and develop models that performs well on day time and night time videos. To tackle these, we explore the spatio-temporal and skeletal fitting methods jointly in a fused system to explore early, late, and combined fusion models to improve overall pedestrian intention prediction on a public data set.

\begin{figure}[ht]
\centerline
{\includegraphics[width=3.2 in, height=1.3 in]{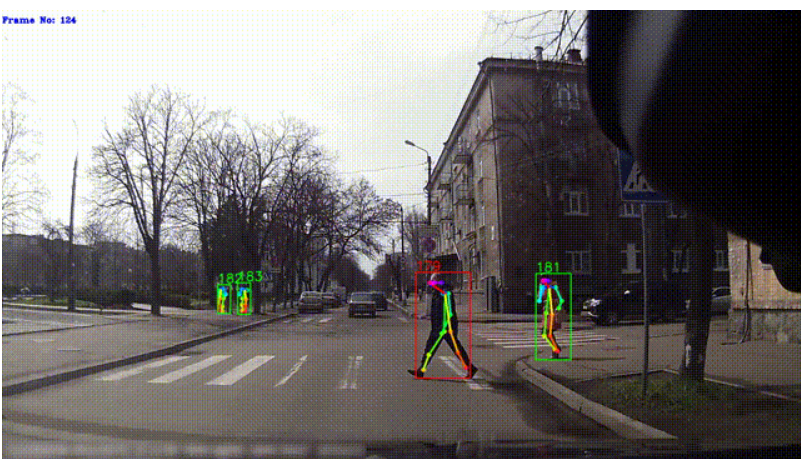}}
\caption{Example of proposed fusion system that combines bounding box and skeletal fitting algorithms for pedestrian intention classification.}
\label{integrate1}
\vspace{-0.4cm}
\end{figure}

In this work, we present such an end-to-end system that is capable of predicting risky intentions of pedestrians up to 16 frames ahead of the actual action, which corresponds to half a second before the \textit{risky maneuver}. 
The main contributions of this work are:
\begin{itemize}
    \item We present a novel pedestrian intention prediction system that relies on fusion from recent state-of-the art methodologies in \cite{saleh2} and \cite{fang}, wherein pedestrian features detected using BBs are combined with pedestrian skeletal features to significantly reduce false positives (see Fig. \ref{integrate1}). 
    \item We describe novel metrics to analyze the accuracy of an end-to-end pedestrian intention prediction/detection system. We present three metrics that capture accuracy of predictions (i) risky crossing behavior of pedestrians up to 16 frames before the motion actually begins, (ii) that motion will continue up to 16 frames in future, and (iii) risky pedestrian motion in up to 16 subsequent frames. 
    \item We evaluate each module of the proposed system with respect to existing works, specifically to analyze model generalizability with respect to input data. In addition, we annotate specific frames from 50 videos from a public dataset \cite{JAAD} to retrain the skeletal fitting module\footnote{Detailed explanation in Supplementary Materials}. These annotations are shared for future benchmarking methods\footnote{Code and demo available at: \url{https://matthew29tang.github.io/pid-model/\#/integrated/}}. 
\end{itemize}

\section{Materials and Methods}
We first describe the  mathematical framework followed by explaining the  methods and data.
\subsection{Mathematical Framework}
Let $\boldsymbol{X}=\{\boldsymbol{x}_{1},\boldsymbol{x}_{2},\ldots\}$
be the sequence of observations per pedestrian and $\boldsymbol{y}=\{y_{1},y_{2},\ldots\}$
be the corresponding intention labels (crossing or not-crossing). The proposed framework is able to learn the latent intention given a new set of observations
$\boldsymbol{X}=\{\boldsymbol{x}_{t}\}_{t=1,2,\ldots T}$, and infer
the labels $\boldsymbol{y}=\{y_{t}\}_{t=1,2,\ldots,T}$. The likelihood
function of the model parameterized by $\boldsymbol{\theta}$ is given
by $p(\boldsymbol{y}\vert\boldsymbol{X},\boldsymbol{\theta}).$

The negative log likelihood $L(\boldsymbol{\theta})$, or loss function, for the training samples
$(\boldsymbol{X}_{i},y_{i}),i=1,2,\ldots n$ can be represented as 
\begin{equation}
  L(\boldsymbol{\theta})=-\sum_{i=1}^{n}\log(y_{i}\vert\boldsymbol{X}_{i},\theta).
\end{equation}
The optimal parameters $\boldsymbol{\theta}^{*}$ for the learned
model can be found as $\boldsymbol{\theta}^{*}=\underset{\boldsymbol{\theta}}{\arg\,\min}\,L\,(\boldsymbol{\theta})$.

Now, for an unseen new observation from the test set $\boldsymbol{x}$,
the most probable label $y^{*}$ will be the one that maximizes the
trained model under the optimized learned parameters $\boldsymbol{\theta}^{*}$ as
\begin{equation}
 y^{*}=\underset{y}{\arg\,\max}\,p({y}\vert\boldsymbol{x},\boldsymbol{\theta}^{*})
 \vspace{-0.3cm}
\end{equation}
In this work we replicate the works in \cite{saleh2} and \cite{fang} to benchmark $\boldsymbol{\theta}^{*}_{\textrm{b}}$ and compare the test performances with the fusion models $\boldsymbol{\theta}^{*}_{\textrm{e}}$, $\boldsymbol{\theta}^{*}_{\textrm{l}}$, $\boldsymbol{\theta}^{*}_{\textrm{c}}$, corresponding to early, late and combination fusion, respectively.

\subsection{Object Detection-based Methods}
The Object Detection-based pedestrian intention framework consists of object detector followed by a object tracker and densetnet classifier. We now describe the modules for pedestrian feature extraction followed by online tracking methods and skeletal fitting algorithms.
\subsubsection{Object Detection Module}
The YOLOv3 (You Only Look Once) algorithm \cite{YOLO} detects 2D bounding boxes around objects of interest (pedestrians). The anchor boxes corresponding to pedestrians with probability greater than 0.5 are returned as bounding boxes (BBs).
\subsubsection{Online Tracking Module}
Once BBs are detected around pedestrians in each image frame, the next step involves tracking each pedestrian across frames with a unique object ID. For this module, we implement two types of online tracking algorithms. The first Simple Online and Realtime Tracking (SORT) algorithm in \cite{saleh2} has one shortcoming that it does not handle occlusions and pedestrians re-entering in a video sequence. Thus, a second tracking algorithm DeepSORT \cite{DeepSORT} is implemented, where pedestrians’ appearance information is used to improve the performance of SORT. This algorithm allows generation of features for person re-identification that can then be compared with the visual appearance of the pedestrians inside the detected bounding boxes to decrease identity switches in \cite{rehder}.
\subsubsection{DenseNet Module} 
\begin{figure}[!ht]
	\centering
	\includegraphics[height=1.0in, width=3.2in]{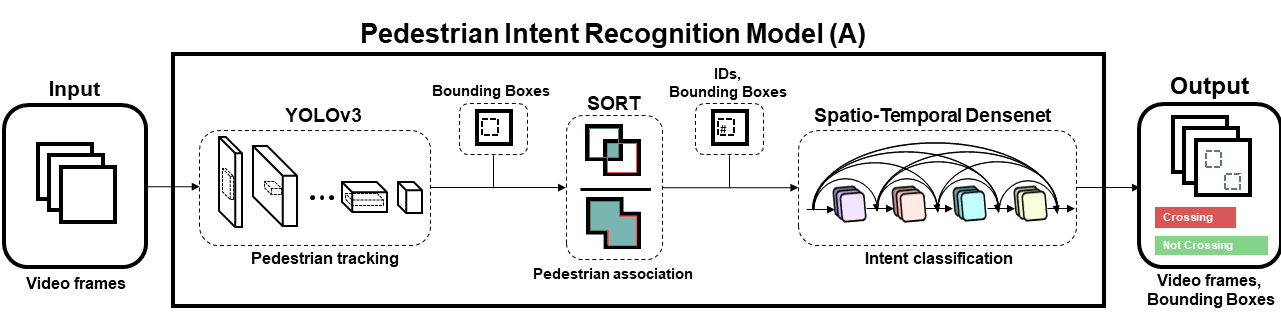}
	\caption{The benchmark pedestrian Intention Prediction Architecture ($\boldsymbol{\theta}^{*}_{\textrm{b}}$). The SORT/DeepSORT modules are interchangeable.  }
	\label{modA}
			\vspace{-0.4cm}
\end{figure}

The next component in our system is a classifier that determines if a pedestrian will cross the street or not. We implement a 121-layer spatio-temporal densenet model \cite{saleh2} and the composite system architecture is shown in Fig. \ref{modA}. 

The densenet in \cite{saleh2} is composed of three dense blocks, where each block comprises of four pairs of $[1\times1\times1]$ and $[3\times3\times3]$ convolutions, respectively. The dense blocks are separated by transition blocks that perform batch normalization, $[1\times1\times1]$ convolutions, and average pooling. All of the model layers are interconnected, which means that the input of layer $l$ is the combination of output from layer $l-1$ and the outputs from each of the previous layers. These connections significantly reduce the number of training parameters because the network can preserve information from prior weights. For the training process, the densenet takes as input a sequence of 16 frames prior to the action of crossing (if applicable), and produces as output two probability scores for crossing or not-crossing for the entire 16 frame sequence. The choice of 16 frames stems from the intent to yield predictions about 0.5 seconds prior to the actual action \cite{saleh2} since camera frames are acquired at 30fps. The integrated system in Fig. \ref{modA} predicts intention by frame, by utilizing a sliding window technique that interpolates frames when the number of frames prior to crossing action is below the minimum requirement (i.e., 16).
\subsection{Skeletal Fitting-based Methods}
Based on \cite{fang}, skeletal fitting models are further applied to bounded pedestrian sub-images to eliminate false positive detections as follows.
\subsubsection{Skeletal-fitting Module}
\begin{figure}[!ht]
	\centering
	\subfigure[Skeletal Fitting Model]
	{\includegraphics[height=2.2cm,width=0.1\textwidth]{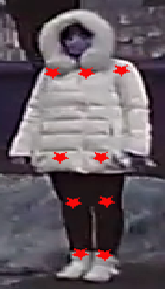}
		\label{ske}}
	\subfigure[Skeletons on 16 frame sequence.]
	{\includegraphics[height=2.2cm,width=0.3\textwidth]{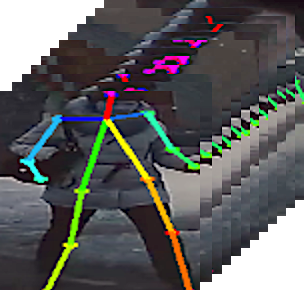}
		\label{ske16}}
	\caption{Skeletal fitting on image sequences.}
			\vspace{-0.4cm}
\end{figure}

The pedestrian BBs from the object detector are used to crop out pedestrian sub-images from the complete image frames. Next, a skeleton fitting algorithm takes the cropped images as input to apply a skeleton onto the pedestrian. The skeleton can contain up to 17 keypoints \cite{cocoann}. Out of these 17 keypoints, 9 are most significant towards pedestrian classification in \cite{fang}. These keypoints consist of the left and right shoulder, hip, knee and ankle, as well as a point between the left and right shoulder as shown in Fig. \ref{ske}. From these 9 keypoints, 396 features based on angles and distances between the skeleton points can be computed for further processing. Examples of the skeleton-fitting process on a sequence of 16 frames is shown in Fig. \ref{ske16}.
\subsubsection{Random Forest (RF) Module}
As an alternative to the densenet model, a RF classifier is implemented for crossing vs not-crossing intent classification. Here, a sliding window method is implemented to extract skeletal features per pedestrian across $t=14$ subsequent frames as in \cite{fang}. Thus, $t\times 396$ features are concatenated per pedestrian followed by RF classification such that the input frames advance by 1 subsequent frame, thereby predicting the intent at the end of 14 frame successions each time.
\subsubsection{Recurrent neural Network (RNN) Module}
Additionally, the RNN model is implemented instead of the RF classifier for intention classification. For this implementation, the frames that did not contain any data are padded with -1 and longer sequences are further divided into shorter versions of maximum 45 frames.
Further, the target data is modified to enable classifier prediction if the pedestrian crosses in the following 14 frames. The input data is normalized to ensure model convergence. The best performing RNN model is a bidirectional LSTM model with one input layer, two hidden layers with 16 memory units each, followed by a dense output layer with 1 memory unit. The  dense output layer uses the sigmoid activation function and the model minimizes (1-classification accuracy) as loss function with Adam optimizer. 
\subsection{Fusion Network Model}
\begin{figure}[ht]
\centerline
{\includegraphics[height=0.8in, width=3.2in]{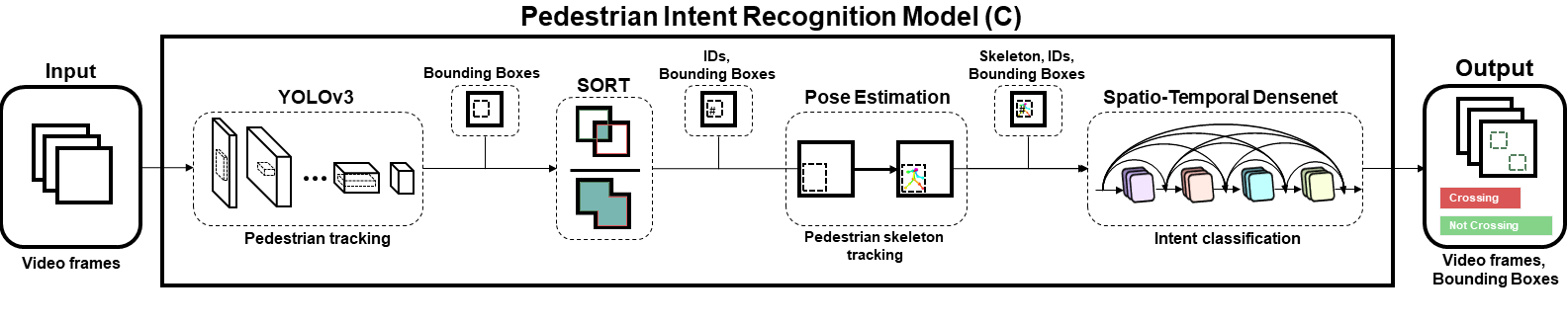}}
\caption{Proposed fusion system architecture that combines bounding box and skeletal fitting algorithms as early fusion ($\boldsymbol{\theta}^{*}_e$) for intention classification.}
\label{integrate}
\vspace{-0.4cm}
\end{figure}

While the bounding box and densenet systems in \cite{saleh2} and skeletal fitting models in \cite{fang} have been analyzed for pedestrian intention detection and prediction, over-detections for crossing action or false positives remain an open problem \cite{varytimidis}. In this work, we fuse the spatial features from BBs with the skeletal features in an early, late and combined fusion setting, with the aim to minimize intention classification false positive errors as shown in Fig. \ref{integrate}.

Here, early fusion consists of fitted skeletons being superimposed on the bounding box regions per pedestrian per image plane to further track and classify intention. Late fusion comprises pre-computed features corresponding to the fitted skeleton to be fed to the last layer of the densenet model as additional features. Thus, 396 features per frame are accumulated over $t$ frames resulting in $t \times 396$ features being combined at the last densenet layer. Combined fusion refers to the combination of early and late fusion setups. We analyze all three setups with respect to the replicated baseline methods in \cite{saleh2} and \cite{fang} to assess the improvement in the overall intention prediction system.
\subsection{Data}
To enable robust system design, large volumes of annotated pedestrian videos are needed that fulfill the following conditions: variations in traffic conditions (urban, parking lots etc.), variations in lighting/weather conditions, benchmarkable performances, public availability, metadata included with annotations (e.g., annotations typically include bounding box coordinates, frame number, crossing vs non-crossing label per pedestrian), additional metadata such as pedestrian age, gaze, posture etc., for pedestrian risk post-processing evaluations.

The following datasets are used in this work for training, analysis and benchmarking modular and overall system performances.
\subsubsection{Joint Attention for Autonomous Driving dataset (JAAD)}
The JAAD \cite{JAAD} is the only publicly available dataset that fulfills all the aforementioned  data requirements. This dataset has been analyzed extensively using bounding box and skeletal fitting models from \cite{saleh2,varytimidis,fang} thus enabling performance benchmarking. In this work, we consider the first 250 videos for model training, while the remaining video sequences from 251 to 346 are considered for the  test dataset. All the benchmarking evaluation is carried out on this dataset.

\subsubsection{Common Object in Context (COCO) Dataset}
Although JAAD includes a variety of pedestrian metadata with regards to appearances, it does not contain skeletal keypoint features which necessitates the use of (COCO) dataset \cite{cocoann}. We use keypoints corresponding to a pedestrians body to train 9-point skeletal fits models. The COCO data set (with 118,000 training images and 5000 test images) is used to generate night time equivalents using the GAN model in \cite{GAN} and the night time images are then used to retrain the skeleton fitting model in \cite{fang}.
 This data set  includes images annotated for object detection, keypoint detection and semantic segmentation. Next, we manually annotate the first 50 videos in JAAD based on the COCO keypoints using the COCO annotator \cite{cocoann}.
 
\subsubsection{Multiple Object Tracking (MOT)}
The MOT challenge is a collection of datasets containing pedestrian annotations. In particular, one of the object trackers in the proposed system is benchmarked on the MOT16 dataset \cite{milanmot16}. 

One limitation of the aforementioned datasets is the lack of night-time videos/sequences. To improve the pedestrian detection/tracking performances for nigh-time sequences as well, we implemented the generative adversarial network (GAN) in \cite{GAN} to process night time equivalents of daytime videos. This data augmentation method enhanced pedestrian detection rates in poor lighting conditions.

\section{Experiments and Results}
Each module of the proposed fusion model is analyzed individually and then in combination to assess their improvements over benchmarked methods. We perform three major experiments in this work. First, we analyze the performances of the object detection, skeletal fitting and classification modules separately with respect to existing benchmarks. Second, we analyze the impact of early, late and combined fusion on intention classification. Third, we analyze the performance of the end-to-end systems with respect to three novel metrics. All modules/models are trained and tested on the same split of the JAAD dataset: the first 250 videos are used for training, and videos from 251 to 346 are used for testing. 

The metrics used to assess detection/prediction performances are average precision (AP), precision (indicative of false positive rate), recall (indicative of false negative rate) and accuracy (acc) as described in \cite{metrics}.
\subsection{Benchmarking Modular Performances}
\subsubsection{Object Detection/Tracking Performance}
We analyze the importance of object detection using the MOT metrics defined in \cite{milanmot16}. In Table \ref{tab:track} the MOT accuracy (MOTA) performances of our implementation of SORT with the annotated groundtruth (GT) is analyzed with respect to the YOLOv3 detections and the VGG16 setup in \cite{varytimidis}. 
\begin{table}[ht]
   \centering
    \caption{Percentage MOTA on Various Sequences.}\label{tab:track} 
    \scalebox{0.7}
    {
    \begin{tabular}{|c c c c c c c c|}\hline
    Method&Overall&TUD&ETH&ETH&ADL&Venice&KITTI\\
    &&Campus&Sunnyday&Pedcross2&Rundle-8&-2&-17\\ \hline
    SORT with GT (Ours)&{\bf 34.0}&36.8&28.6&33.8&{\bf 29.8}&35.4&45.3\\\hline
    SORT with YOLOv3 (Ours)&-&49.4&26.1&38.9&29.0&{\bf36.2}&36.9\\\hline
    SORT with VGG16 \cite{varytimidis}&{\bf 34.0}&{\bf62.7}&{\bf 59.1}&{\bf 45.4}&28.6&18.6&{\bf60.2}\\ \hline
    \end{tabular}}
   \vspace{-0.3cm}
\end{table}

We find that the implementations of SORT with VGG16 in \cite{varytimidis} outperforms the other implementations on most of the reported video sequences. However, the overall MOTA is similar for our implementation and the existing benchmark. 

\subsubsection{Skeletal Fitting Performance}
\begin{figure}[!ht]
	\centering
	\subfigure[Initial Skeletal Fitting]
	{\includegraphics[width=3.3in ,keepaspectratio=True]{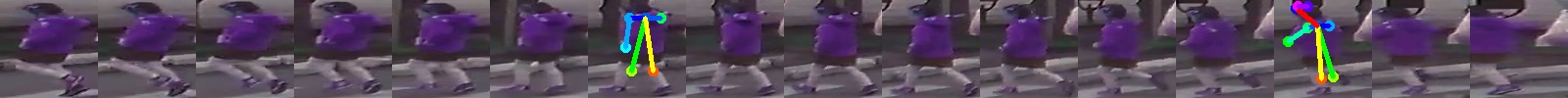}
		\label{skeb}}
	\subfigure[Skeletal fitting retrained on cropped images.]
	{\includegraphics[width=3.3in,keepaspectratio=True]{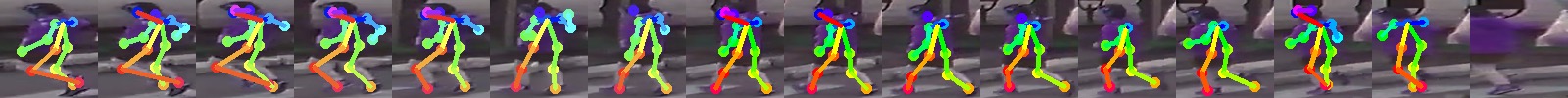}
		\label{skea}}
	\caption{Improvement in skeletal fitting by retraining on JAAD cropped images.}
		
\end{figure}

To benchmark the skeletal fitting module, the first 50 JAAD video sequences are manually annotated for 17 keypoints per pedestrian as described in the supplementary material. The benchmarks are analyzed on a test set that contains 70 sequences made of 16 frames from JAAD in Table \ref{tab:ske}. Three metrics are analyzed here: ratio of found sequences out of total number of sequences ($R_1$), which evaluates the number of sequences out of the 70 test sequences where at least one frame is detected and fitted with a skeleton of atleast 4-keypoints; ratio of found skeletons out of total number of frames ($R_2$), which evaluates the number of skeletons that are found and fitted out of the total $70\times 16$ frames; ratio of found skeletons in found sequences ($R_3$), which evaluates the number of skeletons that are found, in relation to the (found sequences)$\times$16 frames. 
\begin{table}[ht]
   \centering
    \caption{Percentage Performance analysis of Skeletal Fitting.}\label{tab:ske} 
    \scalebox{0.85}
    {
    \begin{tabular}{|c c c c c|}\hline
    Metric&Benchmark \cite{fang}&Retrained on&Retraiend on&Retrained on\\ 
    &&whole images&COCO$+$GAN&cropped images\\ \hline
$R_1$&75.71&30&17.14&\textbf{100}\\\hline
$R_2$&29.11&6.88&4.64&\textbf{82.50}\\\hline
$R_3$&38.44&22.92&27.08&\textbf{82.50}\\ \hline
    \end{tabular}}
    \vspace{-0.4cm}
   \end{table}
In Table \ref{tab:ske}, we observe a significant improvement in skeletal fitting by retraining on cropped images as shown in Fig. \ref{skeb} and Fig. \ref{skea}, respectively.

\subsubsection{Intention Classification Performance}
The pedestrian intention classification performance is analyzed in Table \ref{tab:class}. Here, we are particularly interested in predicting if a pedestrian will cross the road or not a few time frames before to the instant when the action actually begins. For this purpose, we considered a 16 frames interval (around 0.5 seconds in the JAAD dataset) before the frame in which the pedestrian starts crossing according to GT. We observe that the early fusion model results in about 7\% increment in recall and AP over the existing benchmark. The RF and RNN classifiers are evaluated on a subset of the data when compared to the Densenet model as explained in the supplementary material.
\begin{table}[ht]
   \centering
    \caption{Percentage Performance of Classification. }\label{tab:class} 
    \scalebox{0.9}
    {
    \begin{tabular}{|c c c c c|}\hline
Classifier&Features used for 16 frames&AP&Precision&Recall\\\hline
DenseNet \cite{saleh2}&Cropped BBs of Pedestrians&82&74&82\\\hline
DenseNet (Ours)&Early fusion model&\textbf{89}&\textbf{79}&\textbf{89}\\\hline
RF$^{*}$ (Ours)&Skeleton features&81&70&84\\\hline
RNN$^{*}$ (Ours)&Skeleton features&75&73&79\\ \hline
\end{tabular}}
\vspace{-0.4cm}
   \end{table}
\subsection{Fusion Network Model Performance}
The impact of early, late and combination fusion models is analyzed in Table \ref{tab:fusion}. Here, training data is JAAD videos 1-250 and test set are videos 251-346. From Table \ref{tab:fusion} we observe that early fusion is the best approach for the overall system. The primary advantage of early fusion is that it enables pedestrian specific features being extracted by the densenet from the superimposed skeletons on BBs, which leads to lower false positives and higher accuracy. 
\begin{table}[ht]
   \centering
    \caption{Percentage Performance of Fusion Models.}\label{tab:fusion} 
    \scalebox{0.76}
    {
        \begin{tabular}{|c c c c c|}\hline
Metric&Benchmark \cite{saleh2} ($\boldsymbol{\theta}^{*}_{\textrm{b}}$)&Late Fusion ($\boldsymbol{\theta}^{*}_{\textrm{l}}$)&Early Fusion ($\boldsymbol{\theta}^{*}_{\textrm{e}}$)&Combination ($\boldsymbol{\theta}^{*}_{\textrm{c}}$)\\ \hline
Acc&67.5&53.6&\textbf{75.6}&39.0\\ \hline
Loss&0.96&1.94&\textbf{0.93}&1.28\\ \hline
AP&82.8&54.2& \textbf{89.0} &48.5\\ \hline
\end{tabular}}
\vspace{-0.4cm}
   \end{table}

\subsection{End-to-end System Performance Analysis}
Finally, we analyze the performance of the proposed end-to-end system which includes the object detector, tracker and intention classifier. The test data is prepared such that for each crossing pedestrian, their last 30 frames including the frame in which the pedestrian crossed is considered as a test sequence. 

We introduce three novel metrics to assess the performances of the end-to-end models in the moments before and concurrent to the pedestrian crossing action. The three metrics are: $M_1$: the accuracy in predicting the crossing action exactly 16 frames (same as the sliding window used for densenet) before it takes place. This implies prediction regarding crossing intention/or not at the $t-16$ frame regarding an action at $t$ time frame. 
$M_2$: The accuracy in predicting the crossing action in the frame where the action actually takes place. This implies using information from $t-16$ to $t$ time frames to predict a crossing or not crossing action at $t$ time frame. $M_3$: the percentage of ``crossing" or ``not crossing" prediction in the 16 frames proceeding the action. This implies the average accuracy for predicting an action anywhere between the $t-16$ to $t$ frame.
For example, if the GT tells us that the pedestrian is crossing at frame 17, we will check whether our system predicts ``crossing" at frame 1, at frame 17, and the percentage of ``crossing" predictions between frames 1 and 16 using $M_1, M_2, M_3$, respectively. We perform the same procedure for each pedestrian in all the videos to calculate the average percentage accuracy. We exclude the pedestrians whose crossing action happened before the 16th frame, since $M_2, M_3$ are unable to capture this instance.

\begin{table}[ht]
   \centering
    \caption{Percentage Performance Analysis for end-to-end system.}\label{tab:end} 
    \scalebox{0.8}
    {
        \begin{tabular}{|c c c c c|}\hline
Model&Description&$M_1$&$M_2$&$M_3$\\ \hline
A \cite{saleh2}&YOLOv3 $+$ SORT $+$ DenseNet& 37 & \textbf{60} & 55\\ \hline
B&YOLOv3 $+$ DeepSORT $+$ DenseNet & 36 & 30 & 32\\ \hline
C&YOLOv3 $+$ SORT $+$ Early-fused Skeleton $+$ DenseNet&\textbf{45} & 58 & \textbf{57}\\\hline
D&YOLOv3 $+$ DeepSORT $+$ Early-fused Skeleton $+$ DenseNet&40&47&45\\\hline
\end{tabular}}
\vspace{-0.4cm}
   \end{table}
Table \ref{tab:end} shows that model C has the highest accuracy for predicting the crossing intention up to 16 frames ahead with respect to $M_1$ and $M_3$. Thus, when compared with model A and B, the proposed model C can better predict intention with the fusion of skeletons. Although model C is not the highest in $M_2$, its accuracy is very close to the highest. 

\section{Conclusions and Discussion}
In this work we implement multiple fusion models to combine spatio-temporal features with fitted skeletal features to enhance pedestrian intention prediction with respect to state-of-the-art works in \cite{saleh2, fang}. We observe similar to significant improvement in every module with respect to benchmarks owing to additional training on JAAD annotated skeletons on cropped bounding box images. Additionally, we observe that early fusion significantly outperforms late and combination fusion systems. Future works will be directed towards further improving the false negative instances to enable accurate safety distance estimations for ego-vehicle maneuvers.

\bibliographystyle{IEEEtran}
\bibliography{papers}

\end{document}